\documentclass[11pt]{article}
\usepackage{setspace}
\usepackage{emnlp2018}
\usepackage{times}
\usepackage{url}
\usepackage{latexsym}
\usepackage{amssymb}
\usepackage{graphicx}
\usepackage{amsmath}
\usepackage{tipa}
\usepackage{algorithm}
\usepackage[noend]{algorithmic}
\usepackage{cmap}
\usepackage[T1]{fontenc}

\usepackage{hyperref}

\aclfinalcopy

\newcommand{\rerank}{rerank}

\newcommand{\MM}{\textsc{M2M}}
\newcommand{\MP}{\textsc{mpaligner}}

\newcommand{\lm}{language model}
\newcommand{\frequency}{unigram}
\newcommand{\LM}{Language model}

\newcommand{\precision}{precision} %
\newcommand{\cognateTask}{cognate projection} %
\newcommand{\CognateTask}{Cognate projection} %
\newcommand{\DTL}{DTL} %
\newcommand{\Sequitur}{{\normalsize SEQ}}
\newcommand{\RNN}{{\normalsize RNN}}  %
\newcommand{\DTLM}{{\normalsize DTLM}}
\newcommand{\CLUZH}{{\normalsize CLUZH}}
\newcommand{\PlusRR}{{\normalsize DTL+RR}}
\newcommand{\Beinborn}{{\normalsize BZG-13}}

\newcommand{\pluseq}{\mathrel{+}=}

\title{String Transduction with Target Language Models \\
and Insertion Handling}

\author{Garrett Nicolai$^{\dag}$ 
\and Saeed Najafi$^{\ddag}$ 
\and Grzegorz Kondrak$^{\ddag}$ \\
\begin{tabular}{cc}
 & \\
$^{\dag}$Department of Computer Science  & $^{\ddag}$Department of Computing Science\\
Johns Hopkins University                    & University of Alberta\\
{\tt gnicola2@jhu.edu}     &{\tt \{snajafi, gkondrak\}@ualberta.ca}
\end{tabular}
}

\begin{document}
\singlespacing
\maketitle

\begin{abstract}
Many character-level tasks can be framed
as sequence-to-sequence transduction,
where the target is a word from a natural language.
We show that 
leveraging target language models derived from unannotated target corpora,
combined with a precise alignment of the training data,
yields state-of-the art results 
on {\cognateTask}, inflection generation, and phoneme-to-grapheme conversion.
\end{abstract}

\section{Introduction}
\label{sec:intro}

\begin{figure}[t]
\includegraphics[width=80mm,clip=true,trim = 46mm 157mm 44mm 33mm]{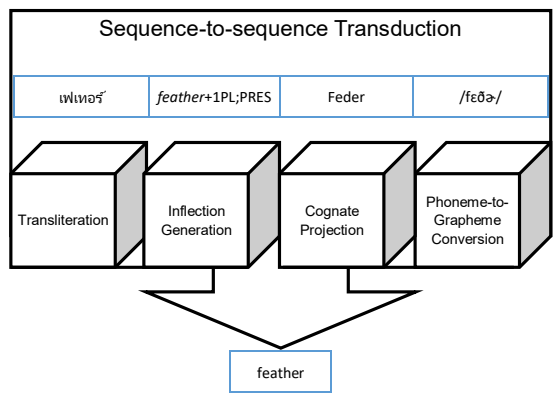}
\vspace{-0.2in} %
\caption{Illustration of four 
character-level sequence-to-sequence prediction tasks.
In each case, the output is a word in the target language.}
\label{fig:subtasks}
\end{figure}

Many natural language tasks, 
particularly those involving character-level operations, 
can be viewed as sequence-to-sequence transduction
(Figure~\ref{fig:subtasks}).
Although these tasks are often addressed in isolation,
they share a common objective ---
in each case, the output is a word in the target language.

The hypothesis that we investigate in this paper is that
a single task- and language-independent system 
can achieve state-of-the-art results 
by leveraging unannotated target language corpora
that contain thousands of valid target word types.
We focus on %
low-data scenarios,
which present a challenge to neural sequence-to-sequence models
because sufficiently large parallel datasets are 
often difficult to obtain.
To reinforce transduction models
trained on modest-sized collections of source-target pairs,
we leverage 
monolingual text corpora 
that are freely available for hundreds of languages.

Our approach is based on discriminative string transduction,
where
a learning algorithm
assigns weights to features
defined on aligned source and target pairs.
At test time,
an input sequence is converted into
the highest-scoring output sequence.
Advantages of discriminative transduction include
an aptitude to derive effective models from small training sets,
as wells as the capability to incorporate diverse sets of features.
Specifically, we build upon 
{\sc DirecTL+} \cite{jiampojamarn2010},
a string transduction tool
which was originally designed for grapheme-to-phoneme conversion.

We present a new system, {\DTLM},
that combines discriminative transduction with
character and word language models (LMs)
derived from large unannotated corpora.
Target language modeling is particularly important
in low-data scenarios,
where the limited
transduction models often produce many ill-formed output candidates.
We avoid the error propagation problem which is inherent in pipeline approaches
by incorporating the LM feature sets directly into the transducer.

In addition,
we bolster the quality of transduction
by employing a novel alignment method,
which we refer to as \emph{{\precision} alignment}.  %
The idea is to allow null substrings ({\em nulls}) on the source side 
during the alignment of the training data,
and then apply a separate aggregation algorithm
to merge them nulls with adjacent non-empty substrings.
This method yields precise many-to-many alignment links 
that lead to improved transduction accuracy. %

The contributions of this paper include the following.
(1) A novel method of 
incorporating strong target language models
directly into discriminative transduction.
(2) A novel approach to unsupervised alignment
 that is particularly beneficial in low-resource settings.
(3) An extensive experimental comparison %
to previous models on multiple tasks and languages,
which includes
state-of-the-art results on inflection generation, cognate projection,
      and phoneme-to-grapheme generation.
(4) Publicly available implementation of the proposed methods.
(5) Three new datasets for {\cognateTask}.

\section{Baseline methods}
\label{sec:baseline}

In this section, 
we describe the baseline methods,
including the alignment of the training data, %
the feature sets of DirecTL+ (henceforth {\DTL}),
and {\rerank}ing as a way of incorporating corpus statistics.

\subsection{Alignment}
\label{sec:alignment}

Before a transduction model can be derived from the training data,
the pairs of source and target strings need to be aligned,
in order to identify atomic substring transformations.
The unsupervised {\MM} aligner \cite{jiampojamarn2007} %
employs the Expectation-Maximization (EM) algorithm with the objective of 
maximizing the joint likelihood of its aligned source and target pairs.
The alignment involves every source and target character.
The pairs of aligned substrings 
may contain multiple characters on both the source and target sides,
yielding {\em many-to-many} (M-M) alignment links.

\begin{figure}[t]
\includegraphics[width=80mm,trim = 34mm 234mm 42mm 20mm]{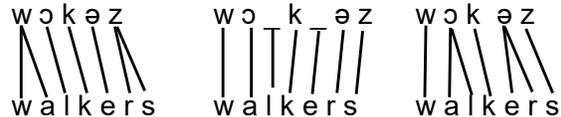}
\caption{Examples of different %
alignments in phoneme-to-letter conversion.
The underscore denotes a null substring.
} %
\label{fig:alignment}
\end{figure}

{\DTL} excludes insertions from its set of edit operations
because they
greatly increase the complexity of the generation process,
to the point of making it computationally intractable \cite{barton:1986:ACL}.
Therefore,
the {\MM} aligner is forced to avoid nulls %
on the source side
by incorporating them into many-to-many %
links 
during the alignment of the training data.
Although many-to-many alignment models are more flexible than 1-1 models,
they also generally require larger parallel datasets
to produce correct alignments.
In low-data scenarios,
especially when the target strings tend to be longer than the source strings,
this approach
often yields sub-optimal alignments 
(e.g., the leftmost alignment in Figure~\ref{fig:alignment}).

\subsection{Features}
\label{sec:features}

{\DTL}  %
is a feature-rich, discriminative character
transducer, which searches for a model-optimal
sequence of character transformation operations %
for its input.
The core of the engine is a dynamic programming
algorithm capable of transducing many consecutive
characters in a single operation, also known
as a semi-Markov model. Using a structured version
of the MIRA algorithm~\cite{McDonald05},
the training process assigns weights to each feature,
in order to achieve maximum separation
of the gold-standard output from all others in the search space.

{\DTL} uses a number of feature templates
to assess the quality of an operation: source context, target
$n$-gram, and joint $n$-gram features. Context
features conjoin the rule with indicators for all
source character $n$-grams within a fixed window
of where the rule is being applied. Target n-grams
provide indicators on target character sequences,
describing the shape of the target as it is being produced,
and may also be conjoined with the source
context features. Joint $n$-grams build indicators
on rule sequences, combining source and target
context, and memorizing frequently-used rule patterns.
An additional copy feature %
generalizes the identity function from source to target,
which is useful if there is an overlap between the input and output symbol sets.

\subsection{Reranking}
\label{sec:RR}

The target language modeling of {\DTL}
is limited to a set of binary $n$-gram features,
which are based exclusively on 
the target sequences from the parallel training data.
This shortcoming can be remedied by 
taking advantage of large unannotated corpora 
that contain thousands of examples of valid target words.
 
\newcite{nicolai2015morph} propose to leverage corpus statistics 
by {\rerank}ing the $n$-best list of candidates 
generated by the transducer. %
They report consistent modest gains 
by applying an SVM-based %
{\rerank}er, %
with features including 
a word unigram corpus presence indicator,
a normalized character language model score, 
and the rank and normalized confidence score generated by {\DTL}.
However, such a pipeline approach suffers from error propagation,
and is unable to produce 
output forms that are not already present in the $n$-best list.
In addition, training a {\rerank}er requires a held-out set
that substantially reduces the amount of training data in low-data scenarios.

\section{Methods}
\label{sec:methods}

In this section, we describe 
our novel extensions:
{\precision} alignment, 
character-level target language modeling, 
and corpus frequency.
We make the new implementation publicly
available.\footnote{http://github.com/GarrettNicolai/DTLM and /M2MP}

\subsection{Precision Alignment}
\label{sec:prec_alignment}

We propose a novel alignment method that produces accurate
many-to-many alignments in two stages.
The first step consists of a standard 1-1 alignment,
with nulls %
allowed on either side of the parallel training data.
The second step removes the undesirable nulls %
on the source side
by merging the corresponding 0-1 links with adjacent 1-1 links.
This alignment approach is superior to 
the one described in Section~\ref{sec:alignment},
especially in low-data scenarios
when there is not enough evidence for many-to-many links.\footnote{The
improvement in the alignment quality 
is relative to the performance of our transduction system,
as we demonstrate in Section~\ref{sec:ablation} ---
the alignments are not necessarily optimal from the linguistic point of view.}

Our {\precision} alignment is essentially a 1-1 alignment 
with 1-M links added when necessary.
In a low-resource setting, an aligner is often unable to distinguish valid
M-M links from spurious ones, 
as both types will have minimal support in the training data.
On the other hand, 
good 1-1 links are much more likely to have been observed. 
By limiting our first pass to 1-1 links, we ensure that 
only good 1-1 links are posited;
otherwise, an insertion is predicted instead.
On the second pass, the aligner
only needs to choose between 
a small number of alternatives for merging the insertions,
increasing the likelihood of a good alignment,
and, subsequently, correct transduction.

Consider the example in Figure~\ref{fig:alignment} where 
5 source phonemes need to be aligned to 7 target letters.
The baseline approach 
incorrectly links the letter `a' with the phoneme /w/
(the leftmost alignment in the diagram).
Our first-pass 1-1 alignment
(in the middle),
correctly matches /\textipa{O}/ to `a', 
while `l' is treated as an insertion.
On the second pass,
our algorithm merges the null %
with the preceding 1-1 link.
By contrast, 
the second insertion, which involves /\textipa{@}/,
is merged with the substitution that follows it
(the rightmost alignment).

\begin{figure}
\begin{center}
\begin{small}
\begin{minipage}{8cm}
	\begin{algorithmic}[1]
	\STATE{\bf{Algorithm:}ForwardInsertionMerging}
	\STATE{\bf{Input:} ($x^{T}$,$y^{V}$)}
	\STATE{\bf{Output:}($\alpha^{T+1,V+1} $)}
	\STATE{$CI=0, PI=0$}
	\FOR{$t$ =0; $t$ $\leq$ $T$}
		\IF{$t > 0$ \AND $x_{t} == \_$}
			\STATE{$CI{\pluseq}1$}
		\ELSE
			\STATE{$PI = CI$}
			\STATE{$CI=0$}
		\ENDIF

		\FOR{$v$=0; $v$ $\leq$ $V$}
			\IF{$t - CI == 0$}
				\STATE{$\alpha_{t,v} = 1$} \STATE{$continue$}
				\COMMENT{insertions at the start of the word}

			\ENDIF
			\IF{$t > 0$ \OR $v > 0$}
				\STATE{$\alpha_{t,v} = 0$} %
			\ENDIF
			\IF{$t > 0$ \AND $v > 0$}
				\FOR{$k =0$; $k$ $\leq$ $PI$}
					\STATE{$\alpha_{t,v}{\pluseq} \, \delta(x_{t-CI-k}^{t}, y_{v-CI-k}^{v}) * $\par$\alpha_{t-CI-k-1,v-CI-k-1}$}
				\ENDFOR
			\ENDIF
		\ENDFOR
	\ENDFOR

\end{algorithmic}
\end{minipage}
\end{small}
\end{center}
\caption{The forward step of {\MM}, modified to merge insertions to adjacent
	 characters.}
\label{fig:pseudocode}
\end{figure}

Figure \ref{fig:pseudocode} demonstrates how we modify the forward
step to merge insertions with adjacent substituions; 
similar modifications are made for
the backward step, expectation, and decoder.
The input consists of a source string {\bf{x}} of length $T$, 
and a target string
{\bf{y}} of length $V$. 
Both {\bf{x}} and {\bf{y}} may contain underscores,
which represent nulls %
from the first alignment pass.
The $\alpha$ score represents the sum of the likelihoods of all paths that
have been traversed through source character $t$ and target character $v$.
In a 1-1 alignment, all $\alpha$ scores accumulate along the diagonal,
while in a many-to-many alignment, other cells of the $\alpha$ matrix
may be filled.  Our {\precision} alignment is a compromise between
these two methods: we consider adjacent characters,
but force the $\alpha$ score to accumulate on the diagonal. 
By allowing insertions and deletions in the first
pass, we force {\bf{x}} and {\bf{y}} to be of equal length.  We then
perform a 1-1 alignment, expanding the alignment size only when the source
character is a null. %

We supplement the forward algorithm of {\MM} with two counters: %
{\rm PI} is the number of adjacent insertions immediately to the left of 
the current character, while {\rm CI} is the number of insertions that
have been encountered since the last substitution.
The loop at line 18
executes the $\alpha$ score accumulation, where $\delta$
is the likelihood of a specific sequence alignment,
effectively merging insertions with adjacent substitutions.
An extended example that illustrates the operation of the algorithm
is included in the Appendix.

\subsection{Character-level language model}
\label{sec:LM}

In order to incorporate a stronger character language model into {\DTL},
we propose an additional set of features 
that directly reflect the probability of the generated subsequences.
We train a character-level language model on a list of types
extracted from a raw corpus in the target language,
applying Witten-Bell smoothing and backoff for unseen $n$-grams. 
During the generation process, 
the transducer incrementally constructs
target sequences character-by-character.
The normalized log-likelihood score of the current output sequence
is computed according to the character language model.

For consistency with other sets of features,
we convert these real-valued scores into binary indicators
by means of binning.
Development experiments led to the creation of bins that
represent a normal distribution around the mean likelihood of words.
Features fire in a cumulative manner, 
and a final feature fires only if no bin threshold is met.
For example, if a sequence has a log-likelihood of \mbox{-0.85}, the
feature for \mbox{-0.9} fires, 
as does the one for -0.975, and \mbox{-1.05}, etc.

\subsection{Corpus frequency counts}
\label{sec:wordFeatures}

We also extend {\DTL} with a feature set that
can be described as a unigram word-level language model.
The objective is to bias the model towards generating
output sequences that correspond to words observed in a large corpus.
Since an output sequence can only be matched against a 
word list after the generation process is complete,
we propose to estimate the final frequency count
for each prefix considered during the generation process.
Following \newcite{cherry2009} we use a prefix trie to
store partial words for reference in the generation phase.
We modify their solution by also storing
the count of each prefix,
calculated as the sum of all of the words in which the prefix occurs.

As with our {\lm} features, {\frequency} features are binned.
A {\frequency} feature fires if the count of the generated sequence
surpasses the bin threshold, in a cumulative manner.

We found that the quality of the target unigram set 
can be greatly improved by 
language-based corpus pruning.
Although unannotated corpora are more readily available
than parallel ones, they are often noisier.
Specifically, crowd-sourced corpora such as Wikipedia
contain many English words that can unduly
influence our {\frequency} features.
In order to mitigate this problem,
we preprocess our corpora
by removing all word unigrams 
that have a higher probability in an English corpus
than in a target-language corpus.

\begin{figure}[t]
\includegraphics[width=80mm,clip=true,trim = 15mm 33mm 20mm 18mm]{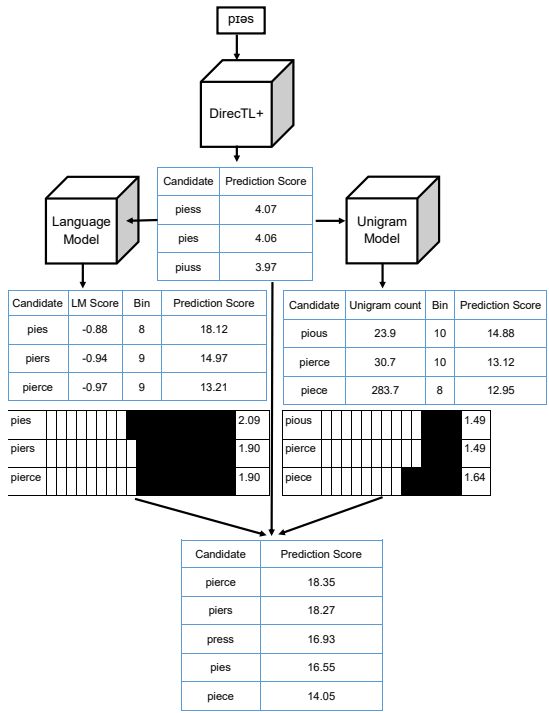}
\vspace{-0.2in} %
\caption{An example of transduction with target language models.
Black cells represent firing features.}
\label{fig:LMProcess}
\end{figure}

Consider an example of how our new features 
benefit a transduction model, 
shown in Figure~\ref{fig:LMProcess}.
Note that although we portray the extensions as part of a pipeline, 
their scores are incorporated jointly with {\DTL}'s other features. 
The top-$n$ list produced by the baseline {\DTL}
for the input phoneme sequence /\textipa{pI@s}/ 
fails to include the correct output {\em pierce}.
However, 
after the new {\lm} features are added, 
the correct form makes its way to the top predictions.
The new features combine with the original features of {\DTL},
so that the high unigram count of {\em piece} 
is not sufficient to make it the top prediction 
on the right side of the diagram.
Only when both sets of new features are incorporated
does the system manage to produce the correct form,
as seen at the bottom of the diagram.

\section{Experiments}
\label{sec:experiments}

In this section, we present the results of our experiments
on four different character-level sequence-to-sequence tasks:
transliteration, 
inflection generation, 
{\cognateTask},
and 
phoneme-to-grapheme conversion (P2G).
In order to demonstrate the generality of our approach,
the experiments involve multiple systems and datasets,
in both low-data and high-data scenarios.

Where low-data resources do not already exist, we simulate a low-data
environment by sampling an existing larger training set.
Low-data training sets consist of 100 training examples, 1000 development
examples, and 1000 held-out examples, except for {\cognateTask}, where
we limit the development set to 100 training examples, and the held-out
set to the remaining examples.
An output is considered correct if it exactly matches any of
the targets in the reference data.

\subsection{Systems}
\label{sec:setup}

We evaluate {\DTLM}, our new system,
against two strong baselines %
and two competitive tools. %
Parameter tuning was performed on the same development sets 
for all systems.

We compare against two baselines.
The first is the standard {\DTL}, 
as described in Section \ref{sec:features}.
The second follows the methodology of \newcite{nicolai2015morph},
augmenting {\DTL} with a {\rerank}er ({\PlusRR}), as described in
Section \ref{sec:RR}.
Both baselines use
the default 2-2 alignment with deletions
produced by the {\MM} aligner.
We train the {\rerank}er using 10-fold cross
validation on the training set, using the {\rerank}ing method
of \newcite{joachims2002optimizing}.
Due to the complexity of its setup on large datasets,
we omit {\PlusRR} in such scenarios.
Except where noted otherwise,
we train 4-gram character  %
language models 
using the CMU %
toolkit\footnote{http://www.speech.cs.cmu.edu/SLM} %
with Witten-Bell smoothing
on the UniMorph corpora of inflected word forms.\footnote{unimorph.org}
Word counts are determined from the first
one million lines of the corresponding Wikipedia dumps.

We also compare against 
Sequitur (\Sequitur),
a generative string transduction tool
based on joint source and target $n$-grams \cite{bisani2008joint},
and a character-level neural model ({\RNN}).
The neural model %
uses the encoder-decoder architecture typically
used for NMT \cite{sutskever2014}.
The encoder is a bi-directional RNN applied 
to randomly initialized character embeddings;
we employ a soft-attention mechanism 
to learn an aligner within the model.
The {\RNN} is trained for a fixed random seed
using the Adam optimizer, 
embeddings of 128 dimensions, and hidden units of size 256.
We use a beam of size 10 to generate the final predictions.
We experimented with the alternative neural approach of
\newcite{makarov2017align}, but found that it only outperforms our {\RNN}
when the source and target sides 
are largely composed of the same set of symbols;
therefore, we only use it for inflection generation.

\subsection{Transliteration}
\label{sec:TL}

Transliteration is the task of converting a word from a source 
to a target script on the basis of the word's pronunciation.

Our low-resource data consists of three back-transliteration pairs
from the 2018 NEWS Shared Task:
Hebrew to English (HeEn), Thai to English (ThEn), and
Persian to English (PeEn).
These languages were chosen because they represent
back-transliteration into English.
Since the original forms were originally English,
they are much more likely to appear in an English corpus
than if the words originated in the source language.
We report the results on the task's 1000-instance development sets.

Since transliteration is mostly used for named entities,
our language model and unigram counts are obtained from a corpus
of named entities.
We query DBPedia\footnote{https://wiki.dbpedia.org}
for a list of proper names, 
discarding names that contain
non-English characters.
The resulting list of 1M names is 
used to train the character language model and
inform the word {\frequency} features.

\begin{table}[t]
\begin{center}
\begin{tabular}{|l||c|c|c|} \hline
System & HeEn & ThEn & PeEn \\ \hline
\hline
{\DTL} & 13.2 & 1.1 & 8.7 \\ %
\hline
{\PlusRR} & 19.0 & 2.7 & 13.6\\ %
\hline
{\DTLM} & \bf{36.7} & \bf{9.6} & \bf{26.1} \\ %
\hline
\hline
{\RNN} & 5.4 & 1.3 & 2.6 \\ \hline
{\Sequitur} & 7.8 & 4.4 & 8.5 \\ %
\hline
\end{tabular}
\end{center}
\caption{Word-level accuracy on transliteration (in \%) 
with 100 training instances.}
\label{tab:transliteration}
\end{table}

\begin{table}[t]
\begin{center}
\begin{tabular}{|l||c|c|c|} \hline
System & HeEn & ThEn & PeEn\\ \hline
\hline
{\DTL} & 21.9 & 37.0 & 23.6 \\
\hline
{\DTLM} & \bf{38.7} & \bf{48.0} & \bf{36.8} \\ %
\hline
\hline
{\RNN} & 25.8 & 43.8 & 26.7 \\ \hline 
{\Sequitur} & 25.5 & 44.9 &  31.2 \\ \hline 
\end{tabular}
\end{center}
\caption{Word-level accuracy on transliteration (in \%) 
with complete training sets.}
\label{tab:transliterationBig}
\end{table}

The results in Table~\ref{tab:transliteration} %
show that 
our proposed extensions 
have a dramatic impact on low-resource transliteration.
In particular,
the seamless incorporation of the target language model
not only simplifies the model but also greatly improves the results
with respect to the {rerank}ing approach.
On the other hand,
the {\RNN} struggles to learn an adequate model
with only 100 training examples.

We also evaluate a larger-data scenario.
Using the same three languages,
we replace the 100 instance training sets with the
official training sets from the 2018 shared task,
which contain 9,447, 27,273, and 15,677 examples for 
HeEn, TnEn, and PeEn, respectively.
The language model and frequency lists are the same
as for the low-resource experiments.
Table \ref{tab:transliterationBig} shows that 
{\DTLM} outperforms the other systems by a large margin
thanks to its ability to leverage a target word list.
Additional results are reported by \newcite{W18-2412}.

\subsection{Inflection generation}
\label{subec:IG}

Inflection generation is the task of producing an inflected word-form,
given a citation form and a set of morphological features.
For example, given the Spanish infinitive {\tt liberar},
with the tag {\tt V;IND;FUT;2;SG}, 
the word-form {\em liberar\'as} should be produced.

In recent years,
inflection generation has attracted much interest 
\cite{dreyer2011discovering,durrett2013supervised,nicolai2015morph,ahlberg2015}.
\newcite{aharoni} propose an RNN augmented with hard attention
and explicit alignments for inflection, but
have difficulty consistently improving upon the
results of {\DTL}, even on larger datasets.
Furthermore, their system cannot be applied to tasks where
the source and target are different languages,
due to shared embeddings between the encoder and decoder.
\newcite{ruzsics} incorporate a language model into the decoder
of a canonical segmentation system.
Our model differs in that we learn the influence 
of the language model during training, 
in conjunction with {\DTL}'s other features.
\newcite{deutsch2018} place a hard constraint on the decoder,
so that it only produces observed derivational forms.
We instead implement a soft constraint, encouraging candidates
that look like real words, but allowing the model to generalize to unseen
forms.

Our inflection data comes from the 2017 %
CoNLL--SIGMORPHON Shared Task on Reinflection~\cite{SIGMORPHON2017}.
We use the datasets from the low-resource setting of 
the inflection generation sub-task,
in which the training sets are composed of 
100 source lemmas with inflection tags
and the corresponding inflected forms.
We supplement the training data with 100 synthetic ``copy'' instances
that simply transform the target string into itself.
This modification,
which is known to help in transduction tasks where the source and target
are nearly identical, %
is used for the inflection generation experiments only.
Along with the training sets from the shared task,
we also use the task's development and test sets, each consisting
of 1000 instances.

Since Sequitur is ill-suited for this type of transduction,
we instead train a model using the method 
of the {\CLUZH} team~\cite{makarov2017align},
a state-of-the-art neural system that was the winner of the 2017 shared task.
Their primary submission in the shared task
was an ensemble of 20 individual systems.
For our experiments,
we selected their best individual system,
as reported in their system paper.
For each language, we train models with 3 separate seeds, and select
the model that achieves the highest accuracy on the development set.

\begin{table}[t]
\begin{center}
\begin{tabular}{|l||c||} \hline
System & Average  \\ \hline
{\DTL} & 40.7 \\ \hline
{\DTLM} & \bf{49.0} \\ \hline
\hline
{\CLUZH} & 40.9 \\ \hline
\end{tabular}
\end{center}
\caption{Word-level accuracy (in \%) on inflection generation
with 100 training instances.}
\label{tab:generation}
\end{table}

Table \ref{tab:generation} shows that {\DTLM} improves upon 
{\CLUZH} by a significant margin.
The Appendix contains the detailed results for individual languages.
{\DTLM} outperforms {\CLUZH} on 46 of the 52 languages.
Even for languages with large morphological inventories,
such as Basque and
Polish, with the sparse corpora that such inventories supply,
we see notable gains over {\DTL}.
We also see large gains for languages 
such as Northern Sami and Navajo
that have relatively small Wikipedias (fewer than 10,000 articles).

{\DTLM} was also evaluated as a non-standard submission in the
low-data track of the 2018 Shared Task on Universal Morphological 
Inflection~\cite{cotterell-conll-sigmorphon2018}.
The results reported by \newcite{UofAST2018}
confirm that 
{\DTLM} substantially outperforms {\DTL} on average.
Furthermore, a linear combination of {\DTLM} and a neural system 
achieved the highest accuracy among all submissions 
on 34 out of 103 tested languages.

\subsection{\CognateTask}
\label{sec:CG}

{\CognateTask}, also referred to as cognate production,
is the task of predicting the spelling of a hypothetical cognate
in another language.
For example, 
given the English word {\em difficulty}, the Spanish word {\em dificultad}
should be produced.  
Previously proposed {\cognateTask} systems have been based on
SVM taggers \cite{mulloni:2007:SRW},
character-level SMT models \cite{beinborn-zesch-gurevych:2013:IJCNLP}, and
sequence labeling combined with a maximum-entropy reranker \cite{ciobanu2016}.

In this section,
we evaluate {\DTLM} in both low- and high-resource settings.
Our low-resource data consists of three diverse language pairs.
The first set 
corresponds to a mother-daughter historical relationship 
between reconstructed Vulgar Latin and Italian (VL-IT)
and contains 601 word pairs 
manually compiled from the textbook of \newcite{boyd1980latin}.
English and German (EN-DE), close siblings from the Germanic family,
are represented by 1013 pairs
extracted from Wiktionary.
From the same source,
we also obtain 438 Slavic word pairs from
Russian and Polish (RU-PL),
which are %
written in different scripts
(Cyrillic vs. Latin).
We make the new datasets publicly available.\footnote{http://github.com/GarrettNicolai/CognateData}

The results are shown in Table~\ref{tab:cognates}.
Of the systems that have no recourse to corpus statistics,
the {\RNN} appears crippled by the small training size,
while {\Sequitur} is competitive with {\DTL}, 
especially on the difficult EN-DE dataset.
On the other hand,
the other two systems %
obtain substantial improvements over {\DTL},
but the gains obtained by {DTLM} 
are 2-3 times greater than those obtained by {\PlusRR}.
This demonstrates the advantage of
incorporating the language model features directly into the training phase
over simple {\rerank}ing.

\begin{table}[t]
\begin{center}
\tabcolsep=0.08cm
\begin{tabular}{|l||c|c|c|} \hline
System & EN-DE & RU-PL & VL-IT \\ \hline
{\DTL} & 4.3 & 23.5 & 39.2  \\ \hline
{\PlusRR} & 7.1 & 32.8 & 43.6 \\  \hline
{\DTLM} & \bf{17.7} & \bf{43.9} & \bf{52.5}\\ \hline
\hline
{\RNN} & 2.2 & 1.7 & 15.7 \\ \hline
{\Sequitur} & 9.2 & 22.3 & 36.9\\ \hline
\end{tabular}
\end{center}
\caption{Word-level accuracy (in \%) on {\cognateTask} 
with 100 training instances.}
\label{tab:cognates}
\end{table}

\begin{table}[t]
\begin{center}
\tabcolsep=0.08cm
\begin{tabular}{|l||c|c|c|} \hline
System & EN-ES & EN-DE & EN-RU \\ \hline
{\Beinborn} & 45.7 & 17.2 & 8.3 \\ \hline
{\DTL} & 30.3 & 24.3 & 13.3 \\ \hline
{\DTLM} &\bf{56.8} & \bf{33.5} & \bf{45.9}\\ \hline \hline
{\RNN}  & 34.3 & 20.5 &  15.0 \\ \hline
\end{tabular}
\end{center}
\caption{Word-level accuracy (in \%) on large-scale {\cognateTask}.
}
\label{tab:cognatesBig}
\end{table}

Our high-resource data %
comes from a previous study of \newcite{beinborn-zesch-gurevych:2013:IJCNLP}.
The datasets were created by applying romanization scripts and string
similarity filters to translation pairs extracted from Bing.
We find that the datasets are noisy, 
consisting mostly of lexical loans from Latin, Greek, and English, 
and include many compound words that share only
a single morpheme (i.e., {\em informatics} and {\em informationswissenschaft}).
In order to alleviate the noise, we found it beneficial to 
disregard all training pairs that could not be aligned by {\MM}
under the default 2-2 link setting.

Another problem in the data is the overlap between the training and test sets,
which ranges from 40\% in EN-ES to 94\% in EN-EL.
Since we believe it would be inappropriate to report results on contaminated
sets,
we decided to ignore all test instances that occur 
in the training data.
(Unfortunately, 
this makes some of the test sets too small for a meaningful evaluation.)
The resulting dataset sizes are included in the Appendix.
Along with the datasets, \newcite{beinborn-zesch-gurevych:2013:IJCNLP}
provide the predictions made by their system.
We re-calculate the accuracy of their predictions (BZG-13), 
discarding any forms
that were present in the training set,
and compare against {\DTL} and {\DTLM}, as well as the {\RNN}.

Table \ref{tab:cognatesBig} shows striking gains.
While {\DTL} and the {\RNN} generally improve upon {\Beinborn},
{\DTLM} vastly outstrips either alternative.
On EN-RU, 
{\DTLM} correctly produces nearly half of potential cognates,
3 times more than any of the other systems.
We conclude that our results constitute a new state of the art
for {\cognateTask}.

\subsection{Phoneme-to-grapheme conversion}
\label{sec:P2G}

Phoneme-to-grapheme (P2G) conversion is the task of 
predicting the spelling of a word from 
a sequence of phonemes that represent its pronunciation
\cite{J96-3003}.
For example, 
a P2G system should convert
[t r ae n z d ah k sh ah n]
into {\em transduction}.
Unlike the opposite task of grapheme-to-phoneme (G2P) conversion,
large target corpora are widely available.
To the best of our knowledge,
the state of the art on P2G
is the joint $n$-gram approach of \newcite{bisani2008joint},
who report improvements on the results of \newcite{galescu2002pronunciation}
on the NetTalk and CMUDict datasets.

Our low-resource dataset consists of three
languages: English (EN), Dutch (NL), and German (DE),
extracted from the CELEX lexical
database~\cite{baayen1995celex2}.

Table \ref{tab:P2G} shows
that our modifications yield
substantial gains for all three languages,
with consistent error reductions of
15-20$\%$ over the {\rerank}ing approach.
Despite only training on 100 words, 
the system is able to convert phonetic transcriptions into
completely correct spellings for a large fraction of words,
even in English, which is notorious for its idiosyncratic orthography.
Once again, the {\RNN} is hampered by the small training size.

We also evaluate {\DTLM} in a large-data scenario.
We attempt to replicate the P2G
experiments reported by \cite{bisani2008joint}.
The data comes from three lexicons
on which we conduct 10-fold cross validation:
English NetTalk~\cite{sejnowski1993nettalk},
French Brulex~\cite{mousty1990brulex},
and
English CMUDict~\cite{weide2005carnegie}.
These corpora contain 20,008, 24,726, and 113,438 words, respectively,
in both orthographic and phonetic notations. 
We note that
CMUDict differs from the other two lexicons in that 
it is much larger, and contains predominantly names,
as well as alternative pronunciations.
When the training data is that abundant, 
there is less to be gained from improving
the alignment or the target language models,
as they are already adequate in the baseline approach.

\begin{table}[t]
\begin{center}
\begin{tabular}{|l||c|c|c|} \hline
System & EN & NL & DE   \\ \hline
{\DTL} & 13.9 & 30.6 & 33.5 \\ \hline
{\PlusRR} & 25.3 & 32.6 & 51.5 \\ \hline
{\DTLM} & \bf{39.6} & \bf{43.7} & \bf{60.5} \\ \hline
\hline
{\RNN} & 2.7 & 6.6 & 14.1 \\ \hline
{\Sequitur} & 15.9 & 30.5 & 28.6 \\ \hline
\end{tabular}
\end{center}
\caption{Word-level accuracy (in \%) on phoneme-to-grapheme conversion
with 100 training instances.}
\label{tab:P2G}
\end{table}

\begin{table}[t]
\begin{center}
\begin{tabular}{|l||c|c|c|} \hline
& NetTalk & Brulex & CMU    \\ \hline
{\DTL} & 61.0 & 68.0 & 48.3 \\ \hline
{\DTLM} & \bf{75.2} & \bf{76.8} & \bf{49.0} \\ \hline
\hline
{\RNN} & 55.8 & 67.1 & 48.0 \\ \hline
{\Sequitur} & 62.7 & 71.5 & 48.6 \\ \hline
\end{tabular}
\end{center}
\caption{Word-level accuracy (in \%) on phoneme-to-grapheme conversion
with large training sets.}
\label{tab:big}
\end{table}

Table \ref{tab:big} shows the comparison of the results.
The P2G results obtained by Sequitur in our experiments
are slightly lower than those reported in the original paper,
which is attributable to differences in data splits, tuned hyper-parameters,
and/or the presence of stress markers in the data.
Sequitur still outperforms the baseline {\DTL},
but {\DTLM} substantially outperforms both Sequitur
and the {\RNN} on 
both the NetTalk and Brulex datasets,
with smaller gains on the much larger CMUDict.
We conclude that our results advance the state of the art on 
phoneme-to-grapheme conversion.

\subsection{Ablation}
\label{sec:ablation}

Table~\ref{tab:ablation} shows the results of disabling 
individual components in the low-resource setting of the P2G task.
The top row reproduces the full {\DTLM} system results
reported in Table \ref{tab:P2G}.
The remaining three show the results
without the character-level LM, word {\frequency},
and {\precision} alignment, respectively.
We observe that the accuracy substantially decreases in almost every case,
which demonstrates the contribution of all three components.

In a separate experiment on the English G2P dataset,
we investigate the impact of the alignment quality
by applying several different alignment approaches to the training sets.
When {\MM} aligner uses unconstrained alignment, it favors
long alignments that are too sparse to learn a transduction model,
achieving less than 1$\%$ accuracy.
\newcite{kubo2011mpaligner}'s {\MP},
which employs a length penalty to discourage such overlong substring matches,
improves moderately, achieving 27.5$\%$ accuracy,
while constraining
{\MM} to 2-2 improves further, to 34.9$\%$.
The accuracy increases to 39.6$\%$ when the {\precision} alignment
is employed.
We conclude that in the low-resource setting,
our proposed {\precision} alignment is preferable to both {\MP}
and the standard {\MM} alignment.

\begin{table}[t]
\begin{center}
\begin{tabular}{|l||c|c|c|} \hline
System & EN & NL & DE   \\ \hline
{\DTLM} & \bf{39.6} & \bf{43.7} & \bf{60.5} \\ \hline
-{\LM} & 37.8 & 38.2 & 48.5 \\ \hline
-Freq & 22.0 & 37.1 & 56.7 \\ \hline
-Precision & 34.9 & 43.7 & 46.7  \\ \hline
\end{tabular}
\end{center}
\caption{Ablation test on P2G data with 100 training instances.}
\label{tab:ablation}
\end{table}

\subsection{Error Analysis}
\label{subsec:EA}

The following error examples 
from three different tasks
demonstrate the advantages of incorporating 
the character-level LM, word frequency,
and {\precision} alignment, respectively.
For the purpose of insightful analysis,
we selected test instances for which
{\DTLM} produces markedly better output than {\DTL}.

In inflection generation,
the second person plural form of {\em knechten}
is correctly predicted as {\em knechtetet}, instead of {\em knechttet}.
In this case, our character language model
derived from a large text corpus
rightly assigns very low probability to the unlikely 4-gram sequence {\tt chtt},
which violates German phonotactic constraints.

In the phoneme-to-grapheme conversion task,
[\textipa{tIlEm@tri}] is transduced to 
{\em t{\textbf{e}}lemetry}, instead of {\em t{\textbf{i}}lemetry}.
In English, a reduced vowel phoneme such as [\textipa{I}]
may correspond to any vowel letter.
In this example,
{\DTLM} is able to successfully leverage 
the occurrence of the correct word-form in a raw corpus.

In {\cognateTask},
the actual cognate of \emph{Kenyan} 
is {\em kenianisch}, rather than {\em kenyisch}.
This prediction can be traced to 
the alignment of the adjectival suffix {\em -an} to {\em -anisch}
in the training data.
The match, which involves a target substring of considerable length,
is achieved through a merger of multiple insertion operations.

The errors made by {\DTLM} fall into a few different categories.
Occasionally, {\DTLM} produces an incorrect form that is more frequent
in the corpus.
For example, {\DTLM} incorrectly guesses 
a subjunctive form of the verb {\emph{ versetzen}}
to be the high-frequency {\em versetzt},
rather than the unseen {\em versetzet}.
More important,
the transducer 
is incapable of generalizing beyond source-target pairs seen in training.
For example, consider 
the doubling of consonants in English orthography 
(e.g. {\em betting}).
Unlike the {\RNN},
{\DTLM} incorrectly predicts the present participle of
{\em rug} as {\em *ruging},
because there is no instance of the doubling of `g' in the training data.

\section{Conclusion}
\label{sec:conclusion}

We have presented {\DTLM}: a powerful language- and task-independent 
transduction system %
that can leverage raw target corpora.
{\DTLM} is particularly effective in low-resource settings,
but is also successful when larger training sets are available.
The results of our experiments
on four varied transduction tasks
show large gains over alternative approaches.

\section*{Acknowledgments}

The first author was supported in part by the Defense Advanced Research
Projects Agency's (DARPA) Low Resource Emergent Incidents (LORELEI) program,
under contract No. HR0011-15-C0113.
Any opinions and conclusions expressed in this material are those 
of the authors, and do not necessarily reflect the views of the Defense
Advanced Research Projects Agency (DARPA).
 
The second and third authors were supported by 
the Natural Sciences and Engineering Research Council of Canada (NSERC).

We thank the members of the University of Alberta teams
who collaborated with us in the context of 
the 2018 shared tasks on transliteration and morphological reinflection:
Bradley Hauer, Rashed Rubby Riyadh, and Leyuan Yu.

\bibliographystyle{acl_natbib.bst}
\bibliography{DirecTLM}

\end{document}

\section*{Appendix}

In this appendix, we present supplemental material
that elaborates on information and results presented in the
paper.

\section{Precision Alignment}

The following figure demonstrates how our modified forward
algorithm calculates the 
$\alpha$ score for this example.
The $\alpha$ score begins at 1.  When we encounter the /w/, there is only one
possible alignment: /w/$\rightarrow$`w', which has likelihood 0.9,
so the $\alpha$ score at
(1,1) is 0.9.  Likewise, the score at (2,2) is the score at (1,1) multiplied
by the only possible alignment likelihood, 0.6, to give 0.54.  At (3,3), we
encounter an insertion, but the only path that goes through (3,3) merges
the insertion to the left.  
The likelihood of /\textopeno/$\rightarrow$ `al' is 0.8.
This is not multiplied by $\alpha$(2,2), but $\alpha$(1,1), which gives 0.72.
$\alpha$(2,2) already contains all paths through `a';
$\alpha$(3,3) is appending `al` to `w', not to `wa'.
$\alpha$(4,4) contains two paths: one that aligns /k/ to `k',
and one that aligns /k/ to `lk'.
Therefore, $\alpha$(4,4) is the sum of 
$\alpha$(3,3) * P(/k/,`k') and $\alpha$(2,2) * P(/k/,`lk'). 
Similarly, $\alpha$(5,5) sums 2 paths: $\alpha$(3,3) * P(/k/,'ke'), and
$\alpha$(2,2) * P(/k/, `lke').
The forward algorithm progresses in this manner until 
it has a score for $\alpha$($T$,$V$).

\begin{figure}[h]
\includegraphics[width=75mm,clip=true,trim = 00mm 20mm 50mm 05mm]{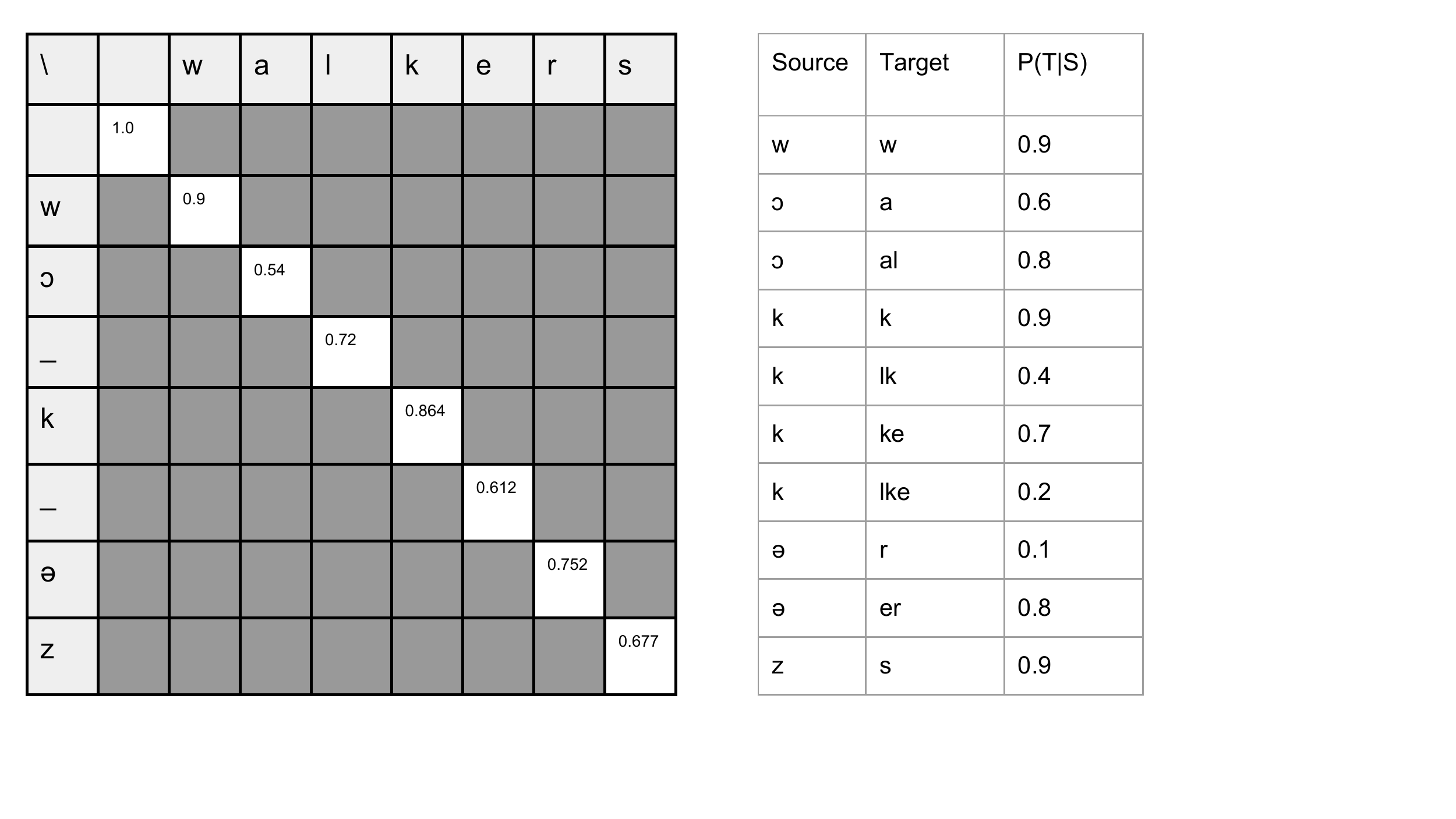}
\label{fig:alphaScore}
\end{figure}

\section{Cognate Generation}

As stated in Section 4.4, the Cognate Generation dataset of
\newcite{beinborn-zesch-gurevych:2013:IJCNLP} 
has substantial overlap between the training and test sets.  
The following table %
provides the sizes of the training and test sets.
Overlap refers to the number of instances in the test set that also occur
in the training data.

\begin{table}[h]
\small
\begin{center}
\begin{tabular}{|c|c|c|c|c|} \hline
Language & Train & Test & Overlap & Our Test \\ \hline \hline
EN-ES & 10531 & 327 & 151 & 176 \\ \hline
EN-DE & 7944 & 1002 & 463 & 539 \\ \hline
EN-RU & 4739 & 127 & 66 & 61 \\ \hline
\end{tabular}
\end{center}
\label{tab:beinborn_data_numbers}
\end{table}

\section{Inflection Generation}

In Section 4.3, we run inflection generation on all 52 languages of the 2017
Shared Task on Morphological Reinflection \cite{SIGMORPHON2017}.
In the paper, we only present the average results, due to space constraints.
The table below %
gives the individual results for each system
and language for the low-data track.\\

\begin{table}[h]
\scriptsize
\begin{center}
\begin{tabular}{|c|c|c||c|} \hline
Language & DTL & DTLM & CLUZH \\ \hline \hline
	Albanian & 16.2 & \bf{21.0} & 9.7 \\ \hline
	Arabic & 21.1 & \bf{33.6} & 20.9 \\ \hline
	Armenian & 47.9 & \bf{56.4} & 50.6 \\ \hline
	Basque & 6.0 & \bf{12.0} & 0.0 \\ \hline
	Bengali & 59.0 & \bf{70.0} & 59.0 \\ \hline
	Bokm\r{a}l & 71.0 & \bf{85.6} & 73.6 \\ \hline
	Bulgarian & 48.8 & \bf{58.6} & 37.5 \\ \hline
	Catalan & 49.2 & \bf{61.9} & 61.8 \\ \hline
	Czech & 36.9 & \bf{41.7} & 34.0 \\ \hline
	Danish & 63.0 & \bf{74.8} & 70.1 \\ \hline
	Dutch & 56.8 & \bf{63.8} & 50.0 \\ \hline
	English & 88.7 & \bf{90.8} & 87.4 \\ \hline
	Estonian & 26.7 & \bf{35.2} & 21.1 \\ \hline
	Faroese & 33.5 & \bf{43.0} & 33.1 \\ \hline
	Finnish & 9.7 & \bf{17.4} & 10.6 \\ \hline
	French & 53.5 & \bf{61.1} & 60.8 \\ \hline
	Georgian & 75.4 & \bf{80.5} & 77.7 \\ \hline
	German & 61.1 & \bf{69.3} & 66.2 \\ \hline
	Haida & 27.0 & \bf{38.0} & 34.0 \\ \hline
	Hebrew & 33.0 & \bf{54.2} & 20.7 \\ \hline
	Hindi & 48.5 & 58.8 & \bf{64.9} \\ \hline
	Hungarian & 27.0 & \bf{39.2} & 30.0 \\ \hline
	Icelandic & 32.6 & \bf{42.1} & 33.6 \\ \hline
	Irish & \bf{32.4} & 29.3 & 18.9 \\ \hline
	Italian & 41.0 & \bf{48.6} & 43.8 \\ \hline
	Khaling & 9.8 & \bf{29.1} & 0.5 \\ \hline
	Kurmanji & 87.0 & \bf{87.1} & 86.1 \\ \hline
	Latin & 17.4 & \bf{27.8} & 13.4 \\ \hline
	Latvian & 59.0 & \bf{62.8} & 62.2 \\ \hline
	Lithuanian & 19.7 & \bf{29.2} & 16.4 \\ \hline
	Lower Sorbian & 46.5 & \bf{51.6} & 47.3\\ \hline
	Macedonian & 51.4 & \bf{68.4} & 54.0 \\ \hline
	Navajo & 14.9 & \bf{19.2} & 7.7 \\ \hline
	Nynorsk & 55.5 & \bf{67.8} & 48.6 \\ \hline
	Persian & 19.2 & 20.0 & \bf{34.6} \\ \hline
	Polish & 46.5 & \bf{48.4} & 38.4 \\ \hline
	Portuguese & 59.9 & 61.1 & \bf{65.5} \\ \hline
	Quechua & 19.3 & 40.3 & \bf{48.2} \\ \hline
	Romanian & 39.5 & \bf{43.7} & 31.4 \\ \hline
	Russian & 41.6 & \bf{50.0} & 36.8 \\ \hline
	Northern Sami & 17.0 & \bf{23.8} & 9.6 \\ \hline
	Scottish & 46.0 & \bf{50.0} & 36.0 \\ \hline
	Serbo-Croatian & \bf{35.8} & 35.5 & 20.6 \\ \hline
	Slovak & 43.8 & \bf{49.6} & 44.9 \\ \hline
	Slovene & 56.5 & 48.4 & \bf{57.6} \\ \hline
	Sorani & 23.6 & \bf{29.9} & 6.5 \\ \hline
	Spanish & 41.6 & \bf{60.2} & 55.5 \\ \hline
	Swedish & 59.7 & \bf{67.2} & 60.1 \\ \hline
	Turkish & 14.7 & 18.4 & \bf{30.8} \\ \hline
	Ukrainian & 44.8 & \bf{60.4} & 37.2 \\ \hline
	Urdu & 37.8 & 55.6 & \bf{67.0} \\ \hline
	Welsh & 44.0 & \bf{54.0} & 39.0 \\ \hline \hline
	Average & 40.7 & \bf{49.0} & 40.9 \\ \hline
\end{tabular}
\end{center}
\label{tab:inflection_results}
\end{table}

\bibliographystyle{acl_natbib.bst}
\nobibliography{DirecTLM}